\documentclass[sigchi]{acmart}

\usepackage{booktabs} 
\usepackage{enumitem}
\usepackage{lipsum}

\setlist[itemize]{leftmargin=*}

\acmConference[MLMH 2018]{2018 KDD workshop on Machine Learning for Medicine and Healthcare}{August 2018}{London, UK}
\acmYear{2018}
\copyrightyear{2018}

\acmPrice{15.00}

\begin{document}
\title[Generating Synthetic but Plausible Healthcare Record Datasets]{Generating Synthetic but Plausible \mbox{Healthcare Record Datasets}} 

\author{Laura Avi\~n\'o}
\authornote{UPC = Universitat Polit\`ecnica de Catalunya}
\orcid{...}
\affiliation{%
  \institution{ESCI-UPF and UPC}
  \state{Barcelona, Spain}
}

\author{Matteo Ruffini}
\orcid{...}
\affiliation{%
  \institution{UPC}
  \state{Barcelona, Spain}
}

\author{Ricard Gavald\`a}
\authornote{BGSMath = Barcelona Graduate School of Mathematics}
\orcid{...}
\affiliation{%
  \institution{UPC and BGSMath}
  \state{Barcelona, Spain}
}

\renewcommand{\shortauthors}{L. Avi\~n\'o et al.}

\begin{abstract}
Generating datasets that ``look like'' given real ones is an interesting tasks for healthcare applications of ML and many other fields of science and engineering. In this paper we propose a new method of general application to binary datasets based on a method for learning the parameters of a latent variable moment that we have previously used for clustering patient datasets. We compare our method with a recent proposal (MedGan) based on generative adversarial methods and find that the synthetic datasets we generate are globally more realistic in at least two senses: real and synthetic instances are harder to tell apart by Random Forests, and the MMD statistic. The most likely explanation is that our method does not suffer from the ``mode collapse'' which is an admitted problem of GANs. Additionally, the generative models we generate are easy to interpret, unlike the rather obscure GANs.Our experiments are performed on two patient datasets containing ICD-9 diagnostic codes: the publicly available MIMIC-III dataset and a dataset containing admissions for congestive heart failure during 7 years at Hospital de Sant Pau in Barcelona.
\end{abstract}

%
%

\ccsdesc[500]{Computing methodologies~Mixture modeling}
\ccsdesc[500]{Computing methodologies~Spectral methods}
\ccsdesc[300]{Applied computing~Health care information systems}

\keywords{Synthetic datasets,
Generative Adversarial Networks,
Method of Moments,
ICD9 codes,
Patient clustering
}

\begin{abstract}
Generating datasets that ``look like'' given real ones is an interesting tasks for healthcare applications of ML and many other fields of science and engineering. In this paper we propose a new method of general application to binary datasets based on a method for learning the parameters of a latent variable moment that we have previously used for clustering patient datasets. We compare our method with a recent proposal (MedGans) based on generative adversarial methods and find that the synthetic datasets we generate are globally more realistic in at least two senses: real and synthetic instances are harder to tell apart by Random Forests, and the MMD statistic. The most likely explanation is that our method does not suffer from the ``mode collapse'' which is an admitted problem of GANs. Additionally, the generative models we generate are easy to interpret, unlike the rather obscure GANs. Our experiments are performed on two patient datasets containing ICD-9 diagnostic codes: the publicly available MIMIC-III dataset and a dataset containing admissions for congestive heart failure during 7 years at Hospital de Sant Pau in Barcelona.
\end{abstract}



\maketitle

\section{Introduction}

The need for synthetic datasets that ``look like'' realistic cases is clear in many fields of science and engineering. For the evaluation of algorithms and software and hardware systems, for example, one wants to establish cross-industry benchmarks against which competitors can be systematically compared. Adequate real datasets may not be available for this purpose, because data owners may not be willing to release them. Also, one may want to stress-test a system by trying it on larger datasets than available, yet be sure that results can be extrapolated to future datasets to come. Finally, synthetic data generators may be used,  by changing synthesis parameters, to controlledly explore situations for which real datasets are not available but might appear in the future. This includes, potentially, research on causal relations in cases in which interventions are not possible. 

In the healthcare domain this need for synthetic but plausible datasets is as acute as anywhere else, because of the high sensitivity of the data. As an example, several healthcare organizations might want to conduct a joint statistical study on their patient populations, 
or investigate in which sense they differ, but 
not willing or able to share their data, even with
a third party. But they could be willing to 
share locally generated synthetic datasets and analyze those -- and of course be cautious at the moment of drawing conclusions. See \cite{PatkiWV16}, for example, for compelling motivation. 

The traditional way of generating synthetic datasets is to create a generative model from scratch using human knowledge, for example a simulator, then run it to generate as many instances as desired. With the progress in machine learning techniques, the first part can be semi-automated by using unsupervised ML techniques to distill the generative model from the real data. 

In recent years, machine learning research on generative models has been boosted by the success of Generative Adversarial Networks (GANs) \cite{goodfellow2014generative} in generating synthetic yet realistic images. This approach, has been latterly used also to the medical domain, for example to generate medical time-series data \cite{gan1}, to simulate individualized treatment effects, reducing in this way the bias of a training dataset \cite{yoon2018ganite}, or to generate realistic synthetic ICD9 records \cite{gan2}.

GANs and their variations have been shown to generate very realistic instances, but present some issues that may limit their applicability to the medical domain. 

First, they suffer of the so-called \textit{mode-collapse} problem \cite{arora2017generalization}, tending to generate with high probability only some of the modes of the underlying distribution they want to emulate. For example, if requested to generate synthetic faces from a dataset of real-world celebrities faces, they will learn a distribution with a domain smaller than expected \cite{arora2017gans}. 
In healthcare this may be problematic: Imagine that we want to generate a synthetic version of the (real) set of patient records for a city. If 90\% of the synthetic patients suffer, say, diabetes, then the generated population is implausible (for the time being, at least!) and useless, even if each synthetic diabetic is plausible because it exhibits the complications and indicators associated in reality to diabetes.

A second major issue of GAN-based approaches is their poor interpretability. Being based on deep generative networks, it is difficult to assess why certain samples are generated with respect to others, and a clear interpretation of the provided generative model is typically missing.

\noindent
{\bf Our contribution.} In this work, similarly to what was done in \cite{gan2},  we focus on the problem of generating synthetic ICD9 records, but instead of relying on a variation of GANs, we use a model-based approach that assumes data to be generated by a certain latent variable model -- precisely, a Naive Bayes model with binary features -- which is learned using the method of moments described in \cite{RuffiniGL17}. In \cite{RuffiniGL17}, the objective of the authors was to learn a model that could be used to cluster the patients appearing in the dataset into groups with similar clinical profiles; here, we leverage on the generative nature of the considered Naive Bayes model, using it to sample realistic synthetic data. The advantages that we expect to obtain with respect to GANs are: 

 \begin{itemize}
\item It is faster to set up and run than GANs and has essentially no hyperparameters to tune, except the number of clusters -- coinciding with the number of latent states of the Naive Bayes model, whose effect is easily determined by evaluation on a test set. 

\item Ideally, modeling the population we are observing as a mixture model with $k$ states should enable to generate not only plausible patient {\em instances} but plausible patient {\em populations}. In fact, the number of latent states can be intuitively used to describe how exhaustively the learned model represents the training population. We would like our synthetic datasets to be harder to distinguish from the real ones than those generated by GANs by relatively general and powerful classifiers as an adversarial might employ. We show this is the case for the MedGAN system described in \cite{gan2}, for Random Forest classifiers, and for two real datasets of patients. We take this as circumstantial evidence that our method does not suffer from mode collapse, or at least to a much smaller degree than MedGan.

\item The generative model synthesized is much easier to interpret than the obscure model generated by GANs. In particular, since it can be described in terms of cluster centers and mixture weights, it is easy to explain to e.g. clinicians in intuitive terms and even visually \cite{RuffiniGL17}.
\end{itemize}

\section{Background}

We assume a dataset $D$ formed by $N$ rows (which we call instances, and sometimes ``patients'') and $d$ columns 
(which we call features, and sometimes ``diagnostics'', ``medications'' etc. depending on the dataset).  
While the method can be extended to continuous values, for simplicity we will assume that attributes are binary, as in the datasets we employ here. We will assume the observed data to be generated by a naive Bayes model with observable binary variables, that is a mixture model characterized by the following generative process:  First a latent (unobservable) variable $Y$ is sampled from discrete distribution: $Y\in \{1,\ldots,k\}$, and $
 \omega_j = P(Y=j)$. The entries of the vector $\omega$ are commonly named the  \textit{mixing weights} of the model.
 Then a vector $X = (X_1,\ldots,X_d)$ of binary observable variables is sampled. Its distribution depends on the value of the latent variable $Y$ and the random variables $X_1,\ldots,X_d$ are conditionally independent given $Y$. It is common to call the entries of $X$ the \textit{features}, whose conditional expectations probabilities are denoted as follows
$P_{i,j} = Prob[X_i=1|Y=j]$.

Taking a modeling perspective, if each instance of our dataset corresponds to a patient, we can imagine the latent variable $Y$ to represent the true, unobservable clinical status of the patient, and the observable features to be the real-world manifestation of this status, which are observable, and depend on the clinical patient state. From a dataset with $N$ samples, the method reported in \cite{RuffiniGL17} enables to 
learn the parameters $((\omega_i)_{i=1..k},(P_{i,j})_{i=1..k,j=1..d})$ of a naive Bayes model accurately describing the data, using a method of moments that is then refined with some steps of Expectation Maximization. 
The description of the method is omitted in this version.

There is no theoretical or experimental reason to believe that this method suffers from mode collapse. In fact, the ``ground truth'' for one of the patient datasets used in \cite{RuffiniGL17} is reasonably known as it contains patients chosen for a particular set of expensive illnesses, and the corresponding profiles appear quite clearly in the retrieved clusters.

\section{Proposed Method}

Consider again the dataset $D$ as a $N\times d$ binary matrix. We aim at generating realistic but synthetic samples ideally indistinguishable from those contained in $D$, following the approach below:
\begin{enumerate}
\item Learn from $D$ a naive Bayes model with $k$ latent states, $((\omega_i)_{i=1..k}(P_{i,j})_{i=1..k,j=1..d})$,
and assume that this model is the one generating the data.
\item Generate $m$ instances following the generative process described above. 
\end{enumerate}
Note that $k$ and $m$ are the only parameters of the process. Naturally, we expect that as $k$ grows the model reflects more accurately the dataset, until it starts to overfit. As the learning technique we employ  to recover from data the naive Bayes model is based on tensor decomposition, we will call our generative approach \textit{TensorGen}. 

We will compare our proposed approach with that of MedGan \cite{gan2}, which are able to generate the same kind of data relying on a traditional GAN enhanced with an auto-encoder allowing the system to generate binary data. This method provided promising results in the original paper but, unlike the method we are proposing here, is not interpretable, and requires the tuning of several hyperparameters.  

A baseline for both methods could be called the
{\em first moment} approximation of the dataset:
Generate each feature independently following its empirical distribution in $D$.  

We consider two methods for comparing a real and a synthetic dataset:

\begin{itemize}
\item Using a relatively powerful predictor builder (we use Random Forests) for the following task: we form a dataset containing equal number of real and synthetic instances. The classifier must, given an instance of this dataset, tell from which side it comes from. For an excellent generator, even the best classifier should only achieve 50\% accuracy at this task.
\item The Maximum Mean Discrepancy metric from \cite{gretton2012kernel}, a statistical test aiming at discriminating whether or not two samples come from the same distribution.
\end{itemize}


Note that we did not use MedGan's discriminator as it would not be a fair comparison, as the generator that MedGan use to sample data is trained exactly with the objective to fool this discriminator.

\section{Dataset(s)}
We will consider two datasets. One will be MIMIC III \citep{johnson2016mimic}, a publicly available dataset containing medical data from the Beth Israel Deaconess Medical Center regarding the years between 2001 and 2012. In particular, we will focus on the diagnostic ICD9 records, a sub-dataset whose rows represent the visits of  patients to the hospital, and the columns contain the codes of the diagnostics  annotated by the doctors. Diagnostics are coded following the well-known ICD9 code \citep{geraci1997international}. The dataset is then mapped into a matrix with binary entries, whose entry $(i,j)$ will be  $1$ if the patient $i$ presents the disease $j$, otherwise a zero.

The second dataset was provided by Hospital de la Santa Creu i Sant Pau in Barcelona\footnote{Unfortunately it cannot be made available because of the confidentiality agreement}. It contains the records of patients admitted to the hospital from 2008 to 2014 because of of congestive heart failure (mostly, ICD9 code 428 as main diagnostic, with a minority of other heart-related codes). Attributes are basic demographics, date and various circumstances of of arrival and discharge, other irrelevant information, and a main diagnostic and up to 9 secondary diagnostics encoded in the ICD9 system. It contains 17250 different rows and about  600 different diagnostics overall. We preprocessed the dataset to keep only the diagnostic features and we kept only the 100 most frequent diagnostics, for which we created 100 binary features.

\section{Results}

The implementation of the method in \cite{RuffiniGL17} is the one available \url{https://github.com/mruffini/NaiveBayesClustering}. 
For Medgan we used the implementation at the following link: 
\url{https://github.com/mp2893/medgan}. 
In each experiment we put reasonable effort at tuning the hyperparameter for best results (though that one can never be sure is one of the method's faults, precisely).
We note that the MedGan implementation uses the GPU if available (it was, in our setting) while our proposed method has not yet been GPU-ized (though it should be possible given that it is mostly performing 
standard linear algebra procedures).

We first describe the experiments on MIMIC-III. As a predictor to tell real and synthetic dataset apart, we tried the sklearn python implementations of Naive Bayes, C4.5 decision trees, Logistic Regression, and Random Forests. Logistic regression and Naive Bayes did not distinguish synthetic and real dataset better than 50\%, even on the baseline; this is to be expected because they are linear methods all three generating methods that we use should get the first-order moments (individual feature probabilities) right. Decision trees did quite well if allowed to grow sufficiently, but not as well as Random Forests. We only report the results using Random Forests from now on. 

Table \ref{mimic} reports various performance statistics for 
the three generating methods. The MMD column is the value of the MMD metric applied to synthetic and real dataset; lower is better (more similar). The other columns indicate the performance of the Random Forest classifier as a distinguisher; unlike most ML papers, {\em worse} predictor performance is {\em better} for our purposes (harder to identify synthetic datasets).

{\small \begin{table}[h]
\centering
\caption{MIMIC}
\label{mimic}
\begin{tabular}{@{}lccccc@{}}
\toprule
             & Accuracy & Recall & Precision & Specificity & MMD  \\ \midrule
Baseline     & 0.86     & 0.84   & 0.88      & 0.89        & 0.12 \\
MedGan       & 0.82     & 0.75   & 0.88      & 0.90        & 0.50 \\
5 clusters   & 0.74     & 0.69   & 0.77      & 0.80        & 0.04 \\
10 clusters  & 0.69     & 0.63   & 0.72      & 0.76        & 0.05 \\
100 clusters & 0.59     & 0.52   & 0.60      & 0.65        & 0.01 \\ \bottomrule
\end{tabular}
\end{table}
}

As can be seen, MedGan performs better than the baseline in the usual metrics, but its MMD is actually quite worse than the baseline; we attribute this to the effect of the mode collapse. Our method performs better than both even if using only 5 latent values (clusters), and gets remarkably low MMD even then. Values improve noticeably up to 100 clusters. 

Table \ref{santpau} shows the results for the second dataset. Again, MedGan does far worse than the baseline in MMD, our method is better on all measures at 10 clusters, and remarkably good at 100 clusters. 

Another aspect to remark is computation time. The experiments with MedGan while our method with $k=10$ clusters takes a few minutes. For $k=100$ the times are in the same order, but one should take into account that MedGan is using the GPU while our method does not. Furthermore, experiments for MedGan have to repeated for a much larger set of hyperparameter value if one wants to be reasonably sure that ``good'' values have been found. 
{\small 
\begin{table}[h]
\centering
\caption{Congestive Heart Failure, Sant Pau}
\label{santpau}
\begin{tabular}{@{}lccccc@{}}
\toprule
             & Accuracy & Recall & Precision & Specificity & MMD  \\ \midrule
Baseline     & 0.72     & 0.68   & 0.75      & 0.77        & 0.59 \\
MedGan       & 0.67     & 0.58   & 0.70      & 0.75        & 3.92 \\
5 clusters   & 0.65     & 0.60   & 0.67      & 0.70        & 0.20 \\
10 clusters  & 0.61     & 0.55   & 0.61      & 0.67        & 0.09 \\
100 clusters & 0.53     & 0.45   & 0.53      & 0.61        & -0.01 \\ \bottomrule
\end{tabular}
\end{table} }

\section{Conclusions and Future Work}

Our method has proven to outperform the GAN-based MedGan on two patient datasets, achieving remarkably low values of MMD and being much harder to distinguish from the real dataset. Besides experimenting on other datasets, future work could include:

\begin{itemize}
\item Investigate in more depth whether the difference in performance of the tested method can be attributed to mode collapse events. 
\item Parallelizing our method to take advantage of GPU.
\item At the theoretical level, investigating whether hard privacy claims can be made about the result of our method, for example in the framework of differential privacy.
\end{itemize}

\section*{Acknowledgements}
We are grateful to the MIMIC project and Hospital de Sant Pau for providing the data, and to Drs. Julianna Ribera, Salvador Benito, and Mireia Puig for their advice.
Research partially supported by grants TIN2017-89244-R and MDM-2014-0445 from MINECO (Ministerio de Economia, Industria y Competitividad) and the recognition 2017SGR-856 (MACDA) from AGAUR (Generalitat de Catalunya).

\bibliographystyle{ACM-Reference-Format}
\bibliography{biblio}

\end{document}